\documentclass[conference]{IEEEtran}
\IEEEoverridecommandlockouts
\usepackage{amsmath,amsfonts}
\usepackage{algorithm}
\usepackage{algpseudocode}

\usepackage{array}
\usepackage{textcomp}
\usepackage{stfloats}
\usepackage{verbatim}
\usepackage{graphicx}
\usepackage{cite}
\usepackage{multirow}
\usepackage{subcaption} 
\usepackage{caption}    
\usepackage{float}
\usepackage{placeins}
\usepackage{wrapfig}
\usepackage[hidelinks,colorlinks=true,linkcolor=black,citecolor=black,urlcolor=black]{hyperref}
\usepackage{makecell}
\usepackage{tikz}

\def\BibTeX{{\rm B\kern-.05em{\sc i\kern-.025em b}\kern-.08em
    T\kern-.1667em\lower.7ex\hbox{E}\kern-.125emX}}

\usepackage{titlesec}
\titlespacing*{\section}{0pt}{*0.8}{*0.8}
\titlespacing*{\subsection}{0pt}{*0.6}{*0.6}
\titlespacing*{\subsubsection}{0pt}{*0.5}{*0.5}

\begin{document}

\title{Adaptive Client Selection in Federated Learning: A Network Anomaly Detection Use Case}
\author{\IEEEauthorblockN{William Marfo, Deepak K. Tosh, Shirley V. Moore} \\
\IEEEauthorblockA{\textit{Department of Computer Science}, \textit{University of Texas at El Paso}, El Paso, USA \\
\ wmarfo@miners.utep.edu, dktosh@utep.edu, svmoore@utep.edu
}}

\maketitle

\begin{abstract}
Federated Learning (FL) has become a ubiquitous approach for training machine learning models on decentralized data, addressing the myriad privacy concerns inherent in traditional centralized methods. However, the efficiency of FL depends on effective client selection and robust privacy preservation mechanisms. Inadequate client selection may lead to suboptimal model performance, while insufficient privacy measures risk exposing sensitive data. This paper proposes a client selection framework for FL that integrates differential privacy and fault tolerance. Our adaptive approach dynamically adjusts the number of selected clients based on model performance and system constraints, ensuring privacy through calibrated noise addition. We evaluate our method on a network anomaly detection use case using the UNSW-NB15 and ROAD datasets. Results show up to a 7\% increase in accuracy and a 25\% reduction in training time compared to FedL2P. Moreover, we highlight the trade-offs between privacy budgets and model performance, with higher privacy budgets reducing noise and improving accuracy. Our fault tolerance mechanism, while causing a slight performance drop, enhances robustness to client failures. Statistical validation using Mann-Whitney U tests confirms the significance of these improvements ($p < 0.05$).
\end{abstract}

\footnote{This material is based upon work supported by the United States Department of Energy’s (DOE) Office of Fossil Energy (FE) Award DE-FE0031744.}


\section{Introduction}
\label{sec:Introduction}

Federated learning (FL) has emerged as a powerful paradigm in machine learning (ML), enabling models to be trained across decentralized data sources without the need to centralize sensitive data \cite{marfo2022network}. This approach is instrumental in privacy-preserving environments, where data cannot be easily shared across borders or entities \cite{trindade2024client}. The efficiency of FL, however, depends heavily on client participation during training. Client selection in FL is not a simple random process but requires careful consideration of various factors such as data heterogeneity, system performance, and computational resources \cite{li2024adafl,trindade2024client,ruan2024valuable}.

The challenges of client selection in FL are multifaceted. First, preserving privacy during client selection and model updates is crucial to prevent adversaries from inferring sensitive information about individual clients' data \cite{yan2023criticalfl}. Second, the unpredictable nature of client availability and potential failures in distributed environments can disrupt the training process, affecting model performance and convergence \cite{yan2023criticalfl}. Additionally, there exists a fundamental trade-off between maintaining strong privacy guarantees and achieving high model accuracy, which complicates the design of effective FL systems \cite{trindade2024client}.

We apply our proposed client selection method to the domain of network anomaly detection, which poses unique challenges due to the complexity and scale of modern networks. Detecting anomalies across distributed and diverse networks, such as the Internet of Things (IoT) systems or automotive communication networks, requires FL techniques that can manage computational efficiency while preserving data privacy. Traditional centralized ML approaches exacerbate privacy concerns, especially when dealing with sensitive network traffic data. To evaluate our method, we chose two prominent datasets: the UNSW-NB15 dataset \cite{moustafa2015unsw}, which captures a range of network traffic patterns and attacks, and the ROAD dataset \cite{verma2024comprehensive}, which focuses on automotive cybersecurity and includes complex, stealthy masquerade attacks. These datasets allow us to test the generalizability of our approach in both typical and specialized network environments, demonstrating its effectiveness in enhancing anomaly detection.

Motivated by these challenges and the limitations of existing approaches, we propose a client selection algorithm that integrates differential privacy (DP) and fault tolerance mechanisms. Our approach combines adaptive client selection with privacy-preserving techniques by applying DP to model updates rather than utility scores. Specifically, we perturb the gradients with Gaussian noise to ensure $(\epsilon, \delta)$-differential privacy, protecting sensitive client data during training \cite{yan2023criticalfl}. The method also includes a robust checkpointing mechanism for fault tolerance, allowing efficient recovery from client dropouts and ensuring continuity in real-world applications. By separating client selection from the privacy-preserving update process, we balance the need for strategic client selection with the necessity of privacy guarantees aligned with FL standards.

The key contributions of this paper are as follows:
\begin{itemize}
    \item We propose a client selection framework that integrates differential privacy through Gaussian noise added to model updates, ensuring privacy preservation without compromising model performance.
    \item We incorporate a fault tolerance mechanism via checkpointing, enhancing system resilience to client failures.
    \item We demonstrate the effectiveness of our approach through experiments on network anomaly detection, achieving significant improvements in accuracy and training efficiency over state-of-the-art baselines.
\end{itemize}

The paper is organized as follows: Section~\ref{sec:Related} reviews related work, Section~\ref{sec:Problem} formulates the problem, Section~\ref{sec:Proposed} presents the client selection method, Section~\ref{sec:Use} outlines the use case and results, and Section~\ref{sec:conclusion} concludes with future work.

\section{Related Work} 
\label{sec:Related}
Trindade et al. \cite{trindade2024client} proposed a two-step client selection method for Hierarchical FL (HFL), achieving notable reductions in resource usage while maintaining model accuracy on datasets like MNIST and CIFAR-10. Despite these improvements, they acknowledge the difficulty of comparing client selection methods due to varied evaluation scenarios and metrics, and their work is based on emulated environments that may not capture real-world complexities. Ruan et al. \cite{ruan2024valuable} examined the trade-offs in FL client recruitment, showing that more clients do not always enhance model performance, particularly in resource-constrained settings. They proposed an optimal solution for client recruitment and highlighted the need to integrate recruitment with existing selection methods. Li et al. \cite{li2024adafl} introduced AdaFL, an adaptive strategy that dynamically adjusts client numbers during training and evaluates contributions based on current and historical performance. AdaFL improved test accuracy and reduced training time. However, the study noted that many strategies still rely on a fixed number of clients, which may not optimally balance training efficiency and model performance throughout the learning process. Huang et al. \cite{huang2023active} proposed ACFL, an active client selection framework for Clustered FL (CFL) with non-IID data, using metrics like uncertainty sampling to select informative clients. Their experiments showed that traditional FL methods, like FedAvg, struggle with class-imbalanced datasets, whereas ACFL significantly improved accuracy and reduced communication overhead. However, they emphasized the need for more refined strategies to effectively handle non-IID data in clustered environments.

Compared to previous studies, this work uniquely integrates adaptive client selection with differential privacy and fault tolerance in FL for network anomaly detection. We introduce an adaptive client selection approach and a checkpointing mechanism using Weibull distribution modeling \cite{vardhan2024modern}, enhancing both performance and fault tolerance. Evaluated on UNSW-NB15 and ROAD datasets, our approach demonstrates improved accuracy and efficiency over baselines like ACFL and FedL2P. This comprehensive solution balances privacy, performance, and fault tolerance, addressing scalability challenges in distributed environments not fully explored in existing literature.

\section{Problem Formulation}
\label{sec:Problem}

We formulate the client selection problem in FL as follows:

Consider a FL system with \( N \) clients, denoted as \( C = \{c_1, c_2, \ldots, c_N\} \), where each client \( c_i \) has a local dataset \( D_i \). The objective is to train a global model \( w \) by aggregating updates from a subset of clients \( S_t \subseteq C \) in each round \( t \), where \( |S_t| = K \). The client selection problem is to find the optimal subset \( S_t^* \) that maximizes performance while satisfying system constraints, including privacy preservation and fault tolerance.

Key assumptions and constraints include non-IID data distribution across clients, privacy preservation through DP (applied to model gradients), limited communication bandwidth, heterogeneous client computational capabilities, and dynamic client availability. We define our objective function as \( F(S_t) = \alpha \cdot \text{Accuracy}(S_t) - \gamma \cdot \text{Cost}(S_t) \), where \( S_t \) is the subset of selected clients in round \( t \), and \( \alpha \) and \( \gamma \) are weighting factors balancing the importance of accuracy and cost, respectively \cite{li2024adafl}. Here, \( \text{Accuracy}(S_t) \) represents the contribution to model performance from selected clients, which could be measured using metrics such as AUC-ROC, as discussed later in the paper. The cost function \( \text{Cost}(S_t) = \sum_{i \in S_t} (\text{Comm}_i + \text{Comp}_i) \) accounts for both communication cost \( \text{Comm}_i \) and computation cost \( \text{Comp}_i \) for client \( i \).  The optimization problem is then to find \( S_t^* = \arg\max_{S_t} F(S_t) \), subject to \( |S_t| = K \) and \( S_t \subseteq \text{Available\_Clients}_t \). The set \( \text{Available\_Clients}_t \) represents the clients that are online and have sufficient resources to participate in the current round, as determined by the GetAvailableClients() function in our algorithm. This formulation incorporates considerations for DP (applied to gradients) and fault tolerance (via checkpointing), which are detailed later in the paper. These mechanisms ensure privacy preservation and system robustness while maintaining the efficiency of the client selection process.

\subsection{Federated Learning Architecture for Adaptive Client Selection}
\begin{figure}[h]
    \centering
    \includegraphics[width=0.6\linewidth]{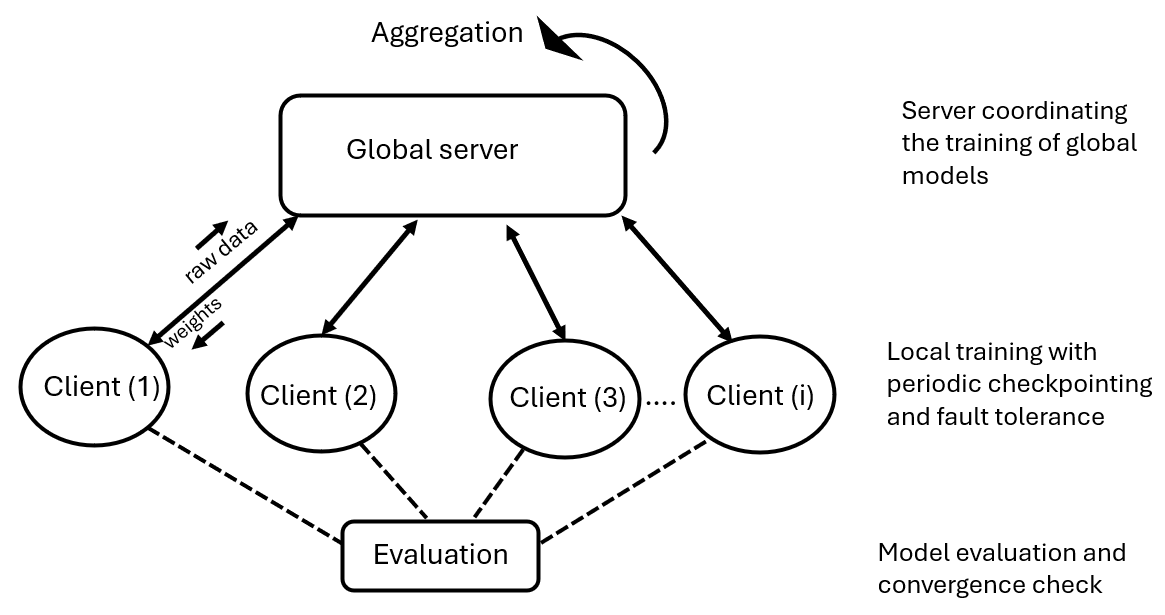}
    \caption{FL architecture showing global server coordination and local client training with checkpointing and fault tolerance.}
    \label{fig:workflow_fl}
\end{figure}

Our FL architecture consists of two main components: \textit{clients} and a \textit{global server}. Fig.~\ref{fig:workflow_fl}  illustrates the FL architecture, highlighting the coordination between the global server and local client training, including checkpointing and fault tolerance mechanisms.

\textbf{1. Clients:} In FL, each client owns local data, which remains on the device to preserve privacy. Each client also maintains a local instance of the model that it trains using its local dataset. Let there be \( N \) clients, denoted as \( c_i \) where \( i \in \{1, \ldots, N\} \). Each client has its dataset \( X_i \), where \( X_i \in \mathbf{R}^{m_i \times d} \) with \( m_i \) representing the number of samples for client \( i \) and \( d \) representing the number of features in each sample.  Each client also has a local model, with parameters denoted by \( w_i \). The local model is trained on the dataset \( X_i \) for a specified number of epochs. After training, the updated parameters \( w_i \) are sent to the global server for aggregation. Specifically, if \( f_i \) represents the function (or architecture) of the local model and \( w_i \) represents the parameters, the training process updates \( w_i \) based on the client’s data \( X_i \). The updated parameters \( w_i \) are then sent to the global server for aggregation in the global model.

\textbf{2. Global Server:} The global server manages the global model by aggregating the updated parameters received from all selected clients. For each round, the global server collects the parameters \( w_i \) from the selected clients and updates the global model parameters \( w_g \) by averaging the received updates. Specifically, if \( N \) clients are selected in round \( t \), the global model parameters are updated as \( w_g = \frac{1}{N} \sum_{i=1}^N w_i \). This iterative process allows the global model to learn from diverse client environments without centralizing sensitive data. After updating the global model, the server broadcasts the updated parameters \( w_g \) to all participating clients for the next round. 

\vspace{-1.4em}\vspace{-1em}

\section{Proposed Client Selection Method}
\label{sec:Proposed}

In this section, we outline our proposed client selection method for FL, which is designed to balance accuracy, privacy, and fault tolerance. We discuss (1) the client selection algorithm, which adaptively selects clients based on their potential contribution to the global model; (2) the implementation of checkpointing strategies, which ensure robust system performance by allowing recovery from client failures; and (3) the integration of DP, which adds noise to client model updates (gradients) to protect against inference attacks.

\subsection{Client Selection Algorithm}
\label{alg:client_selection}
Our selection strategy adapts client participation dynamically based on model performance and system constraints, building upon the approaches in \cite{li2024adafl, yan2023criticalfl}. We compute utility scores for all clients, considering factors like data quality and computational capacity. These scores are solely used for client selection and do not impact user privacy\cite{huang2023active, trindade2024client}. In each communication round, clients are selected based on these utility scores, striking a balance between performance optimization and client diversity. Once selected, they train their local models, and DP is applied by adding Gaussian noise to their model updates (gradients) before sending them to the global server, protecting sensitive data. To ensure fault tolerance, clients periodically checkpoint their progress, enabling recovery in case of failure. After training for a set number of epochs, each client's local model parameters \( w_{f_i} \) are sent to the global server, where they are aggregated to update the global model. Algorithm~\ref{alg:client_selection} summarizes the overall process, incorporating both DP and fault tolerance mechanisms.

\begin{algorithm}
\caption{Client selection with differential privacy and fault tolerance}
\label{alg:client_selection}
\begin{algorithmic}[1]
\Require Set of clients \( C \), clients to select \( K \), privacy budget \( \epsilon \), checkpoint interval \( t_c^* \), failure probability \( p_f \)
\Ensure Selected clients \( S_t \) per communication round \( t \)
\State Initialize utility scores \( U_i \) for all clients \( i \in C \)
\For{each communication round \( t \)}
    \State \( A_t \leftarrow \) GetAvailableClients(\( C \))
    \State \( S_t \leftarrow \) SelectTopK(\( A_t \), \( K \), ComputeUtility(\( U_i \)))
    \For{each client \( i \in S_t \) in parallel}
        \State last\_checkpoint \( \leftarrow \) current\_time()
        \While{not converged}
            \State \( \text{noisy\_grad}_i \leftarrow \text{grad}_i + \mathcal{N}(0, \sigma^2) \)
            \State SendToServer(\( \text{noisy\_grad}_i \))
            \If{current\_time() - last\_checkpoint \( \geq t_c^* \)}
                \State SaveCheckpoint(\( i \))
            \EndIf
            \If{RandomFailure(\( p_f \))}
                \State RecoverFromCheckpoint(\( i \))
            \EndIf
        \EndWhile
    \EndFor
    \State AggregateUpdates(\( S_t \)) and UpdateGlobalModel()
    \If{GlobalModelConverged()}
        \State \textbf{break}
    \EndIf
\EndFor
\State \Return \( S_t \)
\end{algorithmic}
\end{algorithm}

\textbf{Termination condition/epochs in Algorithm~\ref{alg:client_selection}:} 
The algorithm proceeds iteratively over multiple communication rounds, terminating when the global model converges or when the maximum number of rounds is reached. During each round, clients train their local models for a set number of epochs. To ensure fault tolerance, checkpoints are saved at intervals \( t_c^* \), allowing clients to recover from failures without losing progress. If a failure occurs, the client recovers from the last saved checkpoint. This mechanism ensures smooth continuation of the training process while maintaining efficient updates and aggregation at the global server.

\textbf{Handling training failures in decentralized ML:}

To handle client dropouts or disconnections in distributed environments, our framework uses a checkpointing mechanism, enabling smooth recovery and ensuring training continuity \cite{benoit2024checkpointing}.

\paragraph{Recovery protocol without checkpointing}
In the absence of checkpointing, recovery can involve either restarting the training from scratch or reinitializing the failed client’s model using the latest global weights. We prefer the latter as it minimizes interruptions and allows training to continue with minimal disruption, though it may introduce temporary inconsistencies.

\paragraph{Recovery protocol with checkpointing}
In the case of checkpointing, clients periodically store their model state as binary files. When a failure happens, the system restores the client's state from the most recent checkpoint, allowing training to continue without restarting. If a failure occurs during aggregation, the global server can either pause for recovery or reassign the client’s workload to other active clients, ensuring uninterrupted training.

\paragraph{Optimal checkpointing interval}
We estimate the likelihood of client failure using a Weibull distribution, as shown in \cite{vardhan2024modern}, which is well-suited for distributed systems. The failure probability within a checkpointing interval \( t_c \) is modeled as \( p_f(t_c) = 1 - \exp\left(-\left(\frac{t_c}{\lambda}\right)^k\right) \), where \( \lambda \) and \( k \) represent the scale and shape parameters. The total cost function, balancing checkpointing overhead and recovery, is given by \( C(t_c) = \frac{t_c}{T} + p_f(t_c) \cdot \frac{t_r}{T} \), where \( T \) is the total computation time and \( t_r \) is the recovery time. The optimal checkpointing interval \( t_c^* \) is found by solving \( \frac{dC}{dt_c} = 0 \) numerically, using the estimated values of \( \lambda \) and \( k \) derived from historical failure data.

\textbf{Incorporating differential privacy in FL:}

We implement DP in our client selection process to protect client data during communication with the global server. Rather than applying noise to the utility scores, we add Gaussian noise directly to the gradients (model updates) after local training is completed on each client. This ensures that sensitive client data is protected during aggregation at the server. For each selected client \( c_i \), the local gradient \( \nabla w_i \) is perturbed using Gaussian noise to ensure \( (\epsilon, \delta) \)-differential privacy. The perturbed gradient \( \tilde{\nabla w_i} \) is given by \( \tilde{\nabla w_i} = \nabla w_i + \mathcal{N}(0, \sigma^2) \), where \( \mathcal{N}(0, \sigma^2) \) represents Gaussian noise with variance \( \sigma^2 \), calibrated to the privacy budget \( \epsilon \). The noise is added in proportion to the sensitivity of the gradients, ensuring that individual client data remains private. Note that this approach guarantees that, even if an adversary gains access to the global model or intercepts the communication channels, the privacy of individual client data remains protected. We evaluate the impact of different privacy budgets, with lower \( \epsilon \) values offering stronger privacy but potentially reducing model performance due to increased noise. This trade-off between privacy and utility is crucial in FL, as stronger privacy often comes at the cost of reduced accuracy (see results in Section~\ref{sec:Use}-C-2).

\medskip

We now demonstrate the practical application of our client selection method in network anomaly detection, addressing key challenges such privacy preservation, and system robustness.

\section{Use Case: Network Anomaly Detection}
\label{sec:Use}
Network anomaly detection serves as an ideal test case for our client selection framework, given the distributed and privacy-sensitive nature of network traffic data \cite{marfo2022network}. Our approach combines adaptive client selection for diverse network behavior capture, DP for data protection, and fault tolerance for continuous learning despite connectivity issues. We evaluate this framework through experiments (Section~\ref{sec:Experimental}), compare against baselines (Section~\ref{sec:baseline}), and present statistical validation (Section~\ref{sec:Results}).

\subsection{Experimental Setup}
\label{sec:Experimental}
We evaluated our method using the UNSW-NB15 dataset \cite{moustafa2015unsw} for network intrusion detection and the ROAD dataset \cite{verma2024comprehensive} for controller area network traffic, focusing on correlated masquerade attack. Experiments were run on a system with a 12th Gen Intel\textsuperscript{\textregistered} Core\textsuperscript{\texttrademark} i9-12900HK CPU, NVIDIA RTX 3080 Ti GPU, and 32GB RAM, using \texttt{Python 3.8.18}, \texttt{TensorFlow 2.6.0}, and \texttt{PyTorch 0.5.0}. Key hyperparameters ($\epsilon \in [0.1,10.0]$, checkpoint interval $t_c^*$, client fraction $K$) were optimized through grid search over 10 repeated trials with different random seeds, using 200 communication rounds with 5 local epochs per round. Results reported are means across all trials.

\subsection{Baseline Methods and Evaluation Metrics}
\label{sec:baseline}

We compared our client selection method to two baselines: (1) \textbf{ACFL} \cite{yan2023criticalfl}, which selects informative clients using active learning, and (2) \textbf{FedL2P} \cite{lee2024fedl2p}, which applies meta-learning for personalized fine-tuning. We used accuracy and AUC-ROC, standard metrics in anomaly detection, to evaluate performance.

\subsection{Results and Analysis}
\label{sec:Results}
\subsubsection{Detection performance evaluation}
Table \ref{tab:perf_comparison} compares our method to ACFL and FedL2P across 40 clients and 200 communication rounds with 5 local epochs per round, following the neural network model in \cite{marfo2022network}. Our method consistently outperforms the baselines in accuracy, AUC-ROC, and training time due to adaptive client selection, fault tolerance, and differential privacy. On UNSW-NB15, we achieved 94.8\% accuracy in 570 seconds, while on ROAD, we reached 90.3\% accuracy and 0.88 AUC-ROC in 680 seconds.

\begin{figure}[h]
\centering
\includegraphics[width=0.45\textwidth]{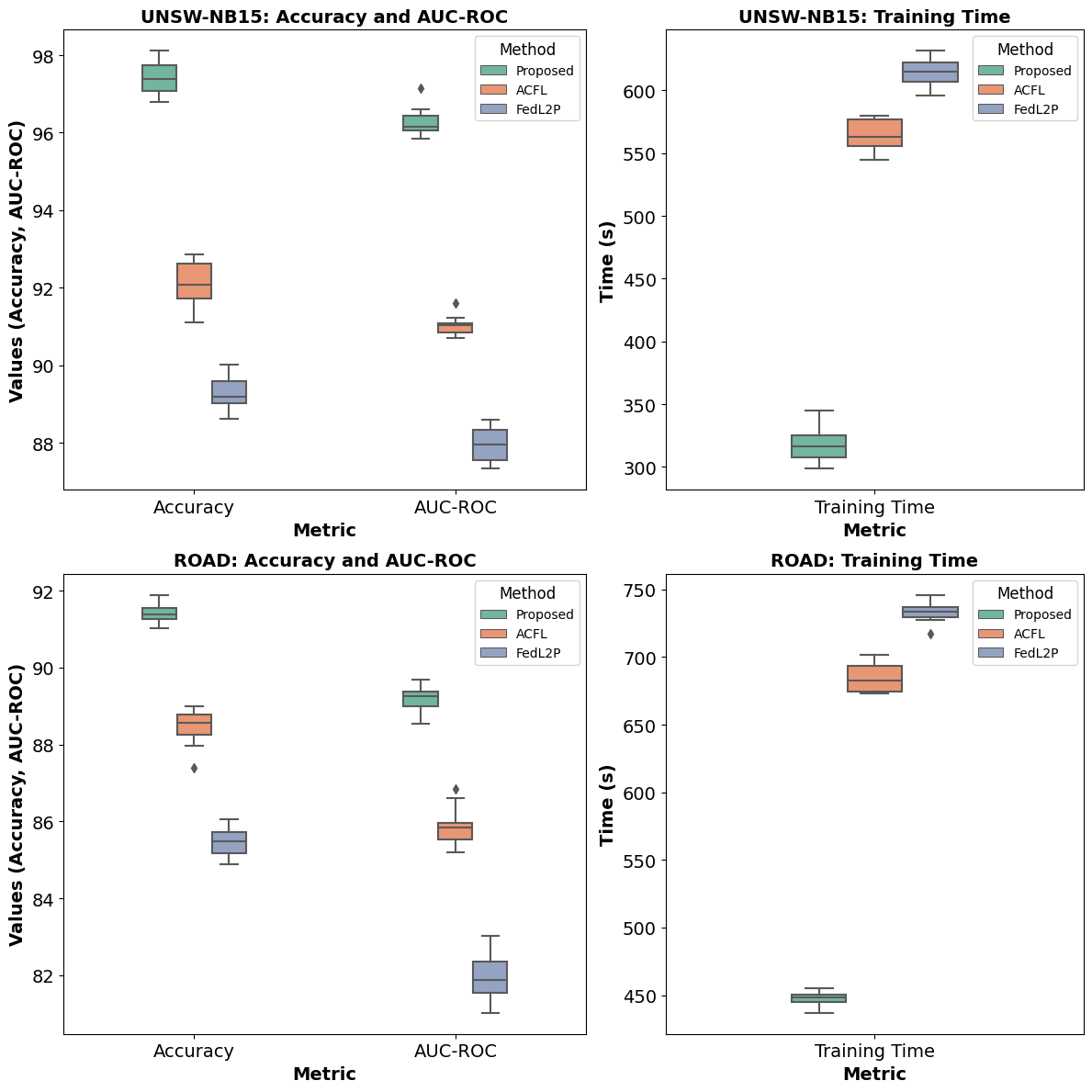}
\caption{Performance comparison of the proposed method, ACFL, and FedL2P across accuracy, AUC-ROC, and training time metrics for the UNSW-NB15 and ROAD datasets.}
\label{fig:performance_comparison}\vspace{-1.4em}
\end{figure}

Fig.~\ref{fig:performance_comparison} illustrates a performance comparison, showing up to 7\% higher accuracy and 25\% faster training compared to baselines. These gains are attributed to the selection of informative clients and balancing privacy with performance, alongside robust fault tolerance, which ensures learning continuity during client failures.

\begin{table}[h]
\centering
\caption{Performance comparison of ACFL, FedL2P, and Proposed method.}
\label{tab:perf_comparison}
\begin{tabular}{l|c|c|c}
\hline
\textbf{Method} & \textbf{Accuracy (\%)} & \textbf{AUC-ROC} & \textbf{Time (s)} \\ \hline
\multicolumn{4}{c}{\textbf{UNSW-NB15}} \\ \hline
ACFL            & 87.8                   & 0.86             & 760              \\
FedL2P          & 92.1                   & 0.91             & 600              \\
Proposed        & \textbf{94.8}          & \textbf{0.93}    & \textbf{570}     \\ \hline
\multicolumn{4}{c}{\textbf{ROAD}} \\ \hline
ACFL            & 83.3                   & 0.81             & 905              \\
FedL2P          & 88.7                   & 0.86             & 710              \\
Proposed        & \textbf{90.3}          & \textbf{0.88}    & \textbf{680}     \\
\hline
\end{tabular}\vspace{-1.4em}
\end{table}

\subsubsection{Impact of differential privacy and fault tolerance}
\begin{figure}[h]
\centering
\includegraphics[width=0.45\textwidth]{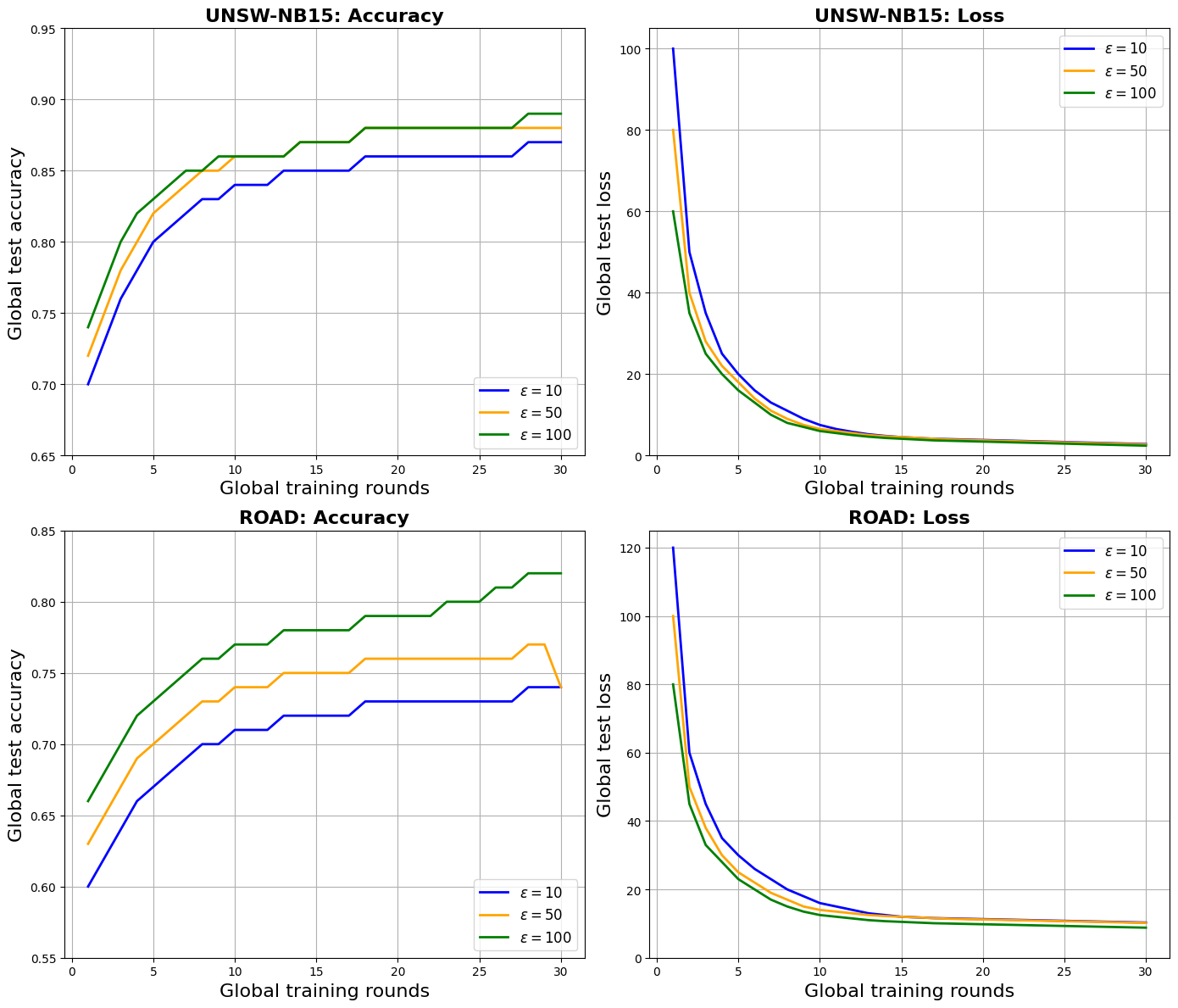}
\caption{Impact of privacy budgets ($\epsilon$ values) on global test accuracy and loss for UNSW-NB15 and ROAD datasets.}
\label{fig:dp_impact}\vspace{-1.4em}
\end{figure}

We evaluate the impact of privacy budgets ($\epsilon$) and fault tolerance on performance. Fig.~\ref{fig:dp_impact} shows that on UNSW-NB15, accuracy improved from 86\% ($\epsilon = 10$) to 89\% ($\epsilon = 100$), with a loss reduction from 3 to 2.5. ROAD showed a greater improvement, with accuracy rising from 73\% to 82\% and loss decreasing from 10 to 9. Table \ref{tab:fault_tolerance} summarizes the impact of fault tolerance. While accuracy and AUC-ROC slightly decrease (e.g., 94.8\% to 92.1\% on UNSW-NB15), fault tolerance enhances robustness, ensuring continuous learning during client failures—essential for real-world scenarios.

Note that trade-offs between communication costs and training time are common in FL \cite{marfo2022network}. Though our approach improves performance, future work will explore the bandwidth-accuracy trade-off in more detail.

\begin{table}[h]
\centering
\caption{Impact of fault tolerance on model performance.}
\label{tab:fault_tolerance}
\begin{tabular}{l|c|c|c}
\hline
\textbf{Configuration} & \textbf{Accuracy (\%)} & \textbf{AUC-ROC} & \textbf{Time (s)} \\ \hline
\multicolumn{4}{c}{\textbf{UNSW-NB15}} \\ \hline
Without fault tolerance & \textbf{94.8} & \textbf{0.93} & 570 \\
With fault tolerance    & 92.1          & 0.91          & \textbf{600} \\ \hline
\multicolumn{4}{c}{\textbf{ROAD}} \\ \hline
Without fault tolerance & \textbf{90.3} & \textbf{0.88} & 680 \\
With fault tolerance    & 88.7          & 0.86          & \textbf{710} \\
\hline
\end{tabular}\vspace{-1.4em}
\end{table}

\subsubsection{Statistical significance testing}
To validate the performance differences between our proposed method and the baselines (ACFL and FedL2P), we employed the Mann-Whitney U test \cite{mann1947test}. We compared AUC-ROC distributions across the UNSW-NB15 and ROAD datasets. Our null hypothesis (\(H_0\)) is that there is no significant difference in performance, while the alternative hypothesis (\(H_1\)) suggests our method's superiority. Table \ref{tab:mann_whitney_results} presents the results, showing consistently low \(p\)-values (\(p < 0.05\)) for all comparisons. Consequently, we reject \(H_0\), confirming that our proposed method significantly outperforms the baselines in terms of AUC-ROC across both datasets.

\begin{table}[h]
\centering
\caption{Mann-Whitney U test results comparing the proposed method against baseline methods.}
\label{tab:mann_whitney_results}
\begin{tabular}{lcc}
\hline
\textbf{Comparison} & \textbf{U Statistic} & \textbf{P-value} \\ \hline
\multicolumn{3}{c}{\textbf{UNSW-NB15}} \\ \hline
Proposed vs. ACFL & 10234.0 & 6.72e-10 \\ 
Proposed vs. FedL2P & 10898.0 & 4.23e-12 \\ \hline
\multicolumn{3}{c}{\textbf{ROAD}} \\ \hline
Proposed vs. ACFL & 9853.0 & 8.45e-08 \\ 
Proposed vs. FedL2P & 10623.0 & 3.97e-11 \\ 
\hline
\end{tabular}\vspace{-1.9em}
\end{table}

\section{Conclusion}
\label{sec:conclusion}
This paper introduced a client selection method for FL with integrated DP and fault tolerance, demonstrating improved performance over FedL2P in network anomaly detection. Our method achieved up to 7\% higher accuracy and 25\% faster training. We identified trade-offs between privacy budgets and model performance, with fault tolerance enhancing robustness at a slight accuracy cost. Limitations include the lack of hyperparameter tuning. Future work will explore adaptive hyperparameter optimization and comparisons with cryptographic techniques, extending applicability to broader domains.

\small
\bibliographystyle{IEEEtran}
\bibliography{70-bibliography}

\vfill

\end{document}